\documentclass{article}

\usepackage{arxiv}

\usepackage[utf8]{inputenc} 
\usepackage[T1]{fontenc}    
\usepackage{hyperref}       
\usepackage{url}            
\usepackage{booktabs}       
\usepackage{amsfonts}       
\usepackage{nicefrac}       
\usepackage{microtype}      
\usepackage{lipsum}
\usepackage{graphicx}
\graphicspath{ {./images/} }
\usepackage{algorithm}
\usepackage{algpseudocode}
\usepackage{tabularx}
\usepackage{microtype}
\usepackage{array}
\usepackage{multirow}
\usepackage{booktabs}
\usepackage{graphicx}
\usepackage[table]{xcolor}
\usepackage[utf8]{inputenc}
\usepackage{xr-hyper}
\usepackage{xcolor}
\hypersetup{filecolor=red}
\usepackage{amsmath}
\usepackage{xspace}
\usepackage{subcaption}
\usepackage[authoryear]{natbib}
\usepackage{natbib}
\usepackage{amsmath}
\usepackage{caption}

\newcommand{\ack}[1]{\section*{Acknowledgements}#1}
\newcommand{\competing}[1]{\section*{Competing Interests}#1}
\newcommand{\contribution}[1]{\section*{Author Contributions}#1}

\title{Field Tracking of Insects Using a Stereoscopic Event-Based Camera Setup}

\author{
 Pratham G. Shenwai \\
  School of Engineering and Technology\\
  University of New South Wales\\
  Canberra, Australia\\
  \And
  Martin J. Lankheet \\
  Experimental Zoology Group\\
  Wageningen University \& Research\\
  Wageningen, The Netherlands.\\
  \And
  John T. Hrynuk \\
  DEVCOM Army Research Lab\\
  Aberdeen Proving Ground, MD, 21005, USA\\
  \And
  Mandiyam Y. Mahadeeswara \\
  Queensland Brain Institute\\
  University of Queensland\\
  St Lucia, QLD, Australia\\
  \And
  Mandyam V. Srinivasan \\
  Queensland Brain Institute\\
  University of Queensland\\
   St Lucia, QLD, Australia\\ 
  \And
  Sridhar Ravi \\
  School of Engineering and Technology\\
  University of New South Wales\\
  Canberra, Australia\\
}

\begin{document}
\maketitle

\begin{abstract}
High-speed tracking of small, fast-moving organisms in their natural environments is important to better understand their behavior and ecology. Traditional frame-based imaging suffers from motion blur due to low temporal resolution, and data storage limitations, propelling a search for more adaptive solutions. Event cameras, which capture changes in brightness at pixel level instead of entire frames, have emerged as a promising solution by increasing temporal resolution and data efficiency. Here, we demonstrate the use of event-based imaging with standard video-based processing methods by converting the asynchronous events into conventional video formats, allowing us to leverage the event camera’s enhanced temporal detail to capture intricate insect flight movements and apply established image analysis techniques. Coupling this conversion process with a stereoscopic configuration provides continuous, low-latency, three-dimensional tracking of fast-moving subjects in field conditions. As a result, we substantially mitigate motion artifacts and achieve more accurate representations of animal movements. By making event-based imaging more readily applicable in natural field settings, our method support broader applications across animal behavior and ecological research, agricultural management, and other fields requiring high-fidelity object tracking in the wild.
\end{abstract}


\section{Introduction}
 Field tracking of insects is a way to understand their behavior, social interaction, movement patterns and their feeding habits. A variety of methodologies have been utilized to identify and track animals in the field including body marking \citep{hagler2001, sendova} , RFID tags \citep{boiteau2001, moreau2011}, Harmonic radars \citep{capaldi2000ontogeny, colpitts2004harmonic}, and camera traps \citep{naqvi2022, droissart2021, straw2011}. However, using camera traps might not be the most effective way to monitor the fast and intricate movements, particularly those of insects as the cameras generally operate at low frame rates and would be subject to motion blur. These limitations of camera traps such as low frame rate and image quality make it a less optimal choice to tracking fast movements. Using high-speed cameras for such applications could serve as an incrementally better alternative \citep{sakakibara2012} however, these cameras are significantly more expensive, and the resulting data is very high in volume with high computational overhead. 
For tracking insects in their natural habitat, a system should be capable of capturing movement in high resolution with minimum data redundancy. Processing large video datasets using computational resources will become a 
bottleneck, limiting the practicality of high-speed cameras for long-duration field studies. An ideal system would thus provide an interface capable of capturing fast movements and managing and reducing data to minimize storage and processing overhead.
 
Event-based cameras offer a promising alternative for field tracking of insects. These cameras are fundamentally different to traditional frame-based imaging by capturing dynamic scenes with greater efficiency \citep{gebauer2024}. Event-based cameras are novel imaging system that have only become commercially available over the last few years. The sensors of event cameras respond only to changes in light intensity, therefore, and each pixel in the sensor array independently detects changes in light intensity. This functionality permits the camera to record information only when there are changes in the visual field, thus reducing data redundancy and improving processing efficiency. Furthermore, the selective sensing results in lower power consumption compared to traditional cameras, making event-based systems ideal for extended deployments in dynamic settings. These cameras also tend to have much higher temporal resolution equivalent to 0.1 microsecond and higher dynamic range which translates to an increased signal to noise ratio even in lower light conditions. 

Here we will present a generalized method for using event-based cameras for tracking flying insects in the field. To demonstrate this, we have chosen honeybees (\textit{Apis} \textit{mellifera}) flying near the entrance of a hive as the object of interest. This was selected because bees approaching and exiting the hive represents a high traffic volume area. This study is one of the first to utilize event-based cameras in a stereo setup to successfully track insects in three-dimensional space, introducing a new diagnostic tool to the community.

\vspace{-0.2cm}
\section{Materials and Methods}

\subsection{Event-Based Camera System }

In this study, we employed a Prophesee EVK4 camera \citep{prophesee2024} to record the insects’ flight trajectories. With a resolution of 1280×720 pixels and a pixel latency under 100 microseconds, this event-based camera can theoretically achieve the temporal equivalent of about 10,000 fps. Unlike traditional frame-based cameras, however, event-based devices produce asynchronous streams of events rather than fixed frames, making direct frame-rate comparisons less meaningful. By accumulating these events over defined time windows, we can create interpretable images that, when sequenced, reveal the insects’ motion patterns.

To extend this approach into three dimensions, we placed two synchronized event-based cameras 40 cm apart, forming a stereoscopic setup that enables triangulation of the insects’ trajectories. Achieving accurate 3D reconstructions requires careful calibration of lens focal lengths, camera alignment, and depth of field. We positioned the system 3.4 m from the hive’s entrance, ensuring that the bees’ approach and departure patterns were clearly captured within the cameras’ field of view.
\begin{figure*}[!ht]
\begin{center}
\includegraphics[height=8cm]{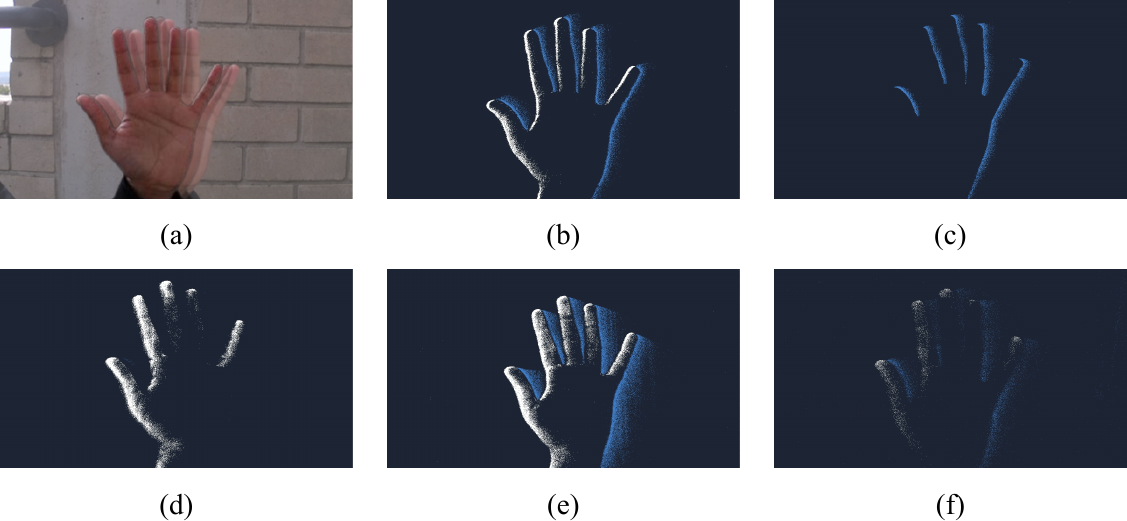}    
 \caption{\textbf{Parameters of an event camera capturing the motion of a palm (moving right to left), with maximum threshold in each case:} (a) Palm recorded by a frame-based camera. (b) Raw frame from an event camera, where white pixels represent ON events and blue pixels represent OFF events. (c) $B_{\text{Con\_ON}}$: Sets the ON contrast threshold, determining how much a pixel's brightness must increase to trigger an ON event. Since the threshold is set to maximum, no ON events are registered. (d) $B_{\text{Con\_OFF}}$: Sets the OFF contrast threshold, determining how much a pixel's brightness must decrease to trigger an OFF event. Since the threshold is set to maximum, no OFF events are registered. (e) $B_{\text{LPF}}$: Adjusts the low-pass filter to smooth out rapid light fluctuations. (f) $B_{\text{HPF}}$: Adjusts the high-pass filter to filter out slow changes in illumination.}
    \label{fig:biassettings}
\end{center}                                 
\end{figure*}

\subsection{Camera Parameter Tuning}

In event-based cameras, the adjustable settings (camera parameters) are designed to handle temporal information. The contrast sensitivity threshold parameters, $B_{{\text{Con\_ON}}}$ and $B_{{\text{Con\_OFF}}}$, adjust how the sensor responds to changes in intensity, setting the threshold levels at which intensity changes are sensed and registered as ON and OFF events respectively (Fig. 1. (c) and (d)). Bandwidth parameter, such as $B_{{\text{LPF}}}$ and $B_{{\text{HPF}}}$, adjust the sensor's filtering capabilities: $B_{{\text{LPF}}}$ minimizes the impact of rapid intensity variations with a low pass filter (Fig.1. (e)), whereas $B_{{\text{HPF}}}$ boosts the camera system's ability to capture dynamic, rapidly changing elements with a high pass filter (Fig.1 (f)), both crucial for accurate sensing. These settings improve the sensor's responsiveness to movement and changes in lighting rather than capturing details of the static regions in the field of view. Thus, it is important that these parameters are tuned appropriately depending on the target being tracked as well as the lighting conditions and background movements that need to be ignored. These parameters can also be adjusted in real-time through either an algorithmic or trial-error approach. 
To demonstrate the workflow we recorded outdoors near a beehive on a clear, sunny day with pervasive sunlight. The lighting conditions were relatively stable, permitting easy equipment tuning without variable shadows or light changes. A higher threshold was maintained for $B_{{\text{Con\_ON}}}$ to minimise sensitivity to sudden increases in brightness due to direct sunlight. This configuration ensured that only profound shifts in the movement of the objects would trigger "ON" events.
The $B_{{\text{Con\_OFF}}}$ setting was kept at a moderate level to address potential minimal fluctuations in light intensity being recognized as "OFF" events.
The setting of $B_{{\text{LPF}}}$ was also changed to filter the slight background noise and ensure stability against minor variations that might affect the accuracy of the data. Please see Supp. Info. for specific details of the tuning parameter values used in our recording sessions.

\subsection{Camera Focusing and Calibration Procedures}
\subsubsection{Camera Output and Focusing}
Frame-based cameras output RGB intensities or greyscale levels for monochromatic setups. With event cameras, the output is asynchronous, consisting of events that are generated when the change in brightness at a pixel exceeds a set threshold. These events include the pixel's location, timestamp, and polarity, where the polarity is 1 for increases in intensity and 0 for decreases. Intensity changes below the threshold do not generate events. These cameras can be equipped with lenses like frame-based imaging to compromise between resolution and field of view. Event cameras also permit adjusting the aperture and focus settings to accommodate various sensing needs. The effects of changing lens parameters, such as focus in event imaging, are similar to frame-based imaging, where de-focusing would activate more pixels, thus fuzzing the image. 

The process of focusing an event camera is different as the cameras only respond to changing intensity. Focusing is carried out systematically to ensure clear and sharp event capture. Initially, the operator chooses a focus target with a flashing pattern positioned at the working distance. To make focusing more straightforward, the aperture is opened to its maximum (with F-number 2), creating a shallow depth of field to enable a sharp and precise focus. The focus distance of the lens is then adjusted until the target appears sharpest. After that, narrowing down the aperture increases the depth of field, expanding the focal range. This procedure establishes a focusing plane where most (or all) honeybees are in crisp focus, thus allowing for the capture of bee trajectories within the focal depth of the image.

\begin{figure*}[!hb]
\begin{center}
\includegraphics[height=6cm]{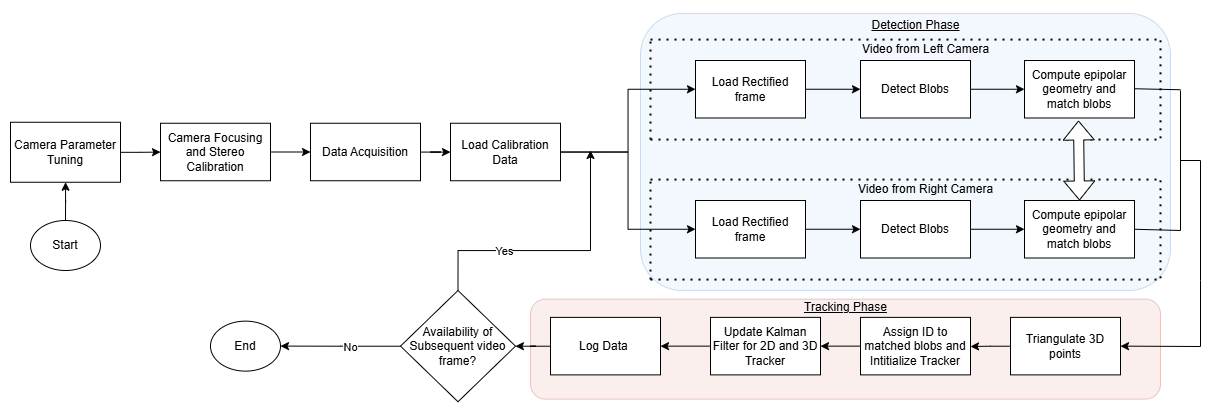}    
\caption{\textbf{Flowchart of the 3D tracking system:} The process begins with camera parameter adjustments, camera focusing, and stereo calibration, followed by data acquisition and loading of calibration data. In the detection phase, rectified frames from both the left and right cameras are processed independently to detect blobs, and epipolar geometry is computed to match the blobs between the two views. In the tracking phase, these matched blobs are first used to triangulate 3D points, and then each set of points is assigned an ID. A Kalman filter is subsequently updated to track the objects in both 2D and 3D space. Throughout this process, data from each tracker is continuously logged to facilitate 3D trajectory plotting.}  
\label{fig2}                                 
\end{center}                                 
\end{figure*}

\subsubsection{Stereo Calibration}
Stereo calibration is a pivotal step in setting up the camera systems for depth estimation and three-dimensional reconstructions. The process of stereo calibration for event cameras is nominally similar to standard image based imaging systems with some modifications. Stereo calibration is performed once the cameras are individually focused at the expected working distance. The cameras are then mounted to a rigid setup, and their orientations are adjusted to have overlapping fields of view. An optical-geometric model is determined for both cameras with parameters specifying their relative positions and orientations. The process consists of aligning the cameras to a standard coordinate system to triangulate points in the observed scene. These are usually computed by analyzing a calibration pattern- a standard checkerboard- from numerous perspectives using both cameras \citep{zhang2000camera}. A flashing checkerboard pattern is used as a tool for calibration here, considering the features of the cameras used in our study (Fig. 3. (a)). A dynamic method was chosen here to provide more visibility and contrast for the calibration points. This intermittent flashing defines the edges of checks such that the check identification and alignment during calibration is much easier. In this study, 80 checkerboard images were captured from each camera throughout the volume, in various orientations, by moving the checkerboard to improve spatial accuracy.
As established in standard stereo vision methodologies \citep{ZouLing}, precise calibration is essential for ensuring accurate epipolar geometry, facilitating correct pixel correspondences, and ultimately achieving reliable 3D reconstruction.

\subsection{Data Acquisition and Processing}
 In this work, the accumulation time used to generate video files was 10,000 microseconds. This resulted in an effective frame rate of 100 fps, which is well-suited for capturing the fine details in insect motions. We recorded the flight of honeybees from a distance of 3.4m from the hive (Fig. 3(b)). 
 We converted the event file (.raw format to a video format (.avi) to apply conventional image processing techniques. In our data acquisition process, the activities of bees were recorded over several 40-second sessions, with careful parameter adjustment to ensure a high signal-to-noise ratio.

\subsection{Detection and Tracking Algorithm}
Existing OpenCV-based techniques \citep{bradski2000} were used for detecting and tracking insects, honeybees in this case (Fig. 2). The general approach consists of combining greyscaling, noise filtering, thresholding, erosion, dilation, contouring of the image followed by applying a Kalman filter to robustly identify and track individual bees in the field of view. This approach uses well-established algorithms for reliable results in dynamic complex settings. This study examines the algorithm in two interdependent parts: the detection and tracking phases. Although the methods were initially designed for frame-based video, they are applied to video generated from an event camera by first converting the sparse, asynchronous event data into a frame-based format. This conversion allows conventional methods such as morphological operations (erosion, dilation) and contouring to identify the honeybees in the field of view. Once the bees are detected, a Kalman filter tracks their movement across the generated video frames. This hybrid approach leverages the high temporal resolution and reduced motion blur inherent in event-based cameras while utilizing well-established image processing techniques to detect and track the bees. The main advantage is that, despite converting to frame-based video, the underlying event camera data provides superior tracking accuracy in challenging conditions, such as fast movements, low light, and high-density scenes.

\begin{figure*}[!hb]
\begin{center}
\includegraphics[height=11cm]{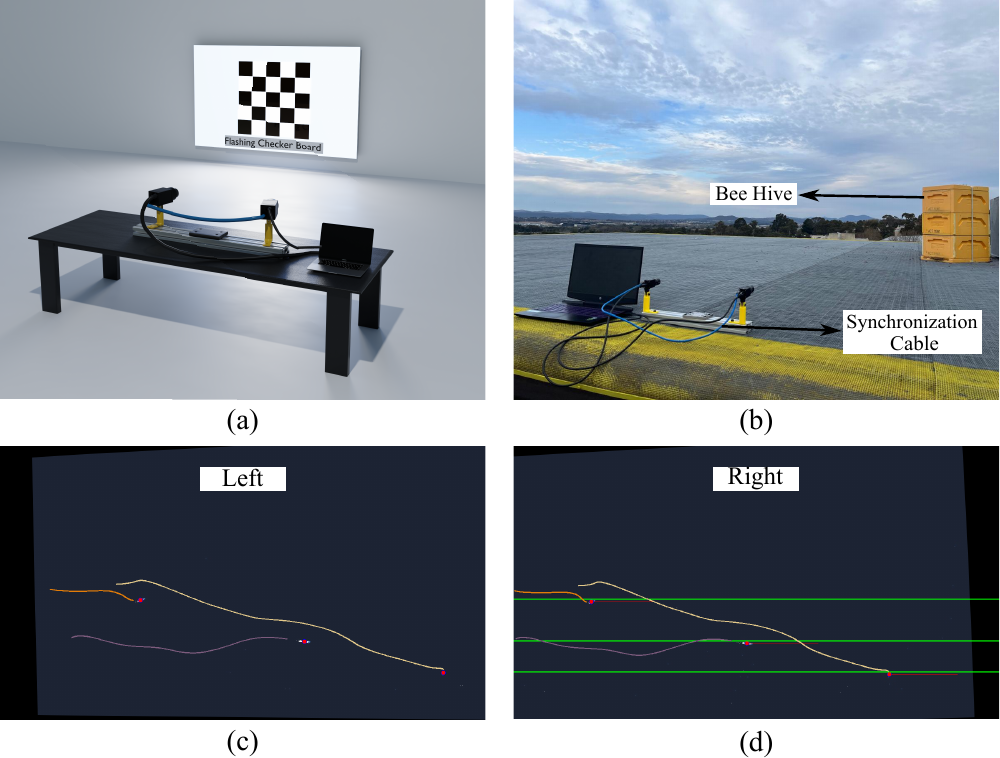}    
\caption{\textbf{Stereo Calibration and Bee Tracking Using Epipolar Geometry:} (a) Stereo calibration using a flashing checkerboard to obtain intrinsic and extrinsic parameters. (b) Data acquisition by observing the hive from a distance of approximately 3.4 meters using a stereo setup of event camera, with a blue synchronization cable connecting both cameras. (c) 
Rectified frame from the left camera detecting and tracking 3 bees in the scene. (d) A rectified frame from the right camera shows detections with epipolar lines overlaid. A red dot on the bee confirms a successful match across both frames. Each colored line represents the 2D trajectory of an individual bee. Once matched, each bee is assigned a consistent ID, represented by the same color used for its trajectory across all frames.}  
\label{fig:fig3}                                 
\end{center}                                 
\end{figure*}

\subsubsection{Detection Phase}
During the detection phase, the images are pre-processed by converting them to greyscale. Due to the asynchronous nature of the data, singular pixel noise may be present from the recording sessions. To eliminate this noise, we apply median blurring, thresholding and morphological operations such as erosion and dilation. Additionally, contour detection techniques are used to accurately outline the honey bees, as shown in Figure S2.
Median blur is a technique used to reduce noise in an image, particularly singular pixel noise, by replacing each pixel's value with the median value of its neighboring pixels. This operation smooths out noise while preserving edges, and the resulting pixel values typically remain within the range of 0 to 255 for greyscale images. However, to uniquely identify subjects or regions of interest in the image, a binary thresholding step is often applied. Binary thresholding converts the image into a binary format, where pixel values are set to either 0 (black) or 255 (white) based on a chosen intensity threshold. This ensures that the subjects or regions of interest are represented distinctly with white pixels (255) against a black (0) background, improving the clarity of segmentation and feature identification. Binary thresholding remains essential for event camera data processing because it ensures that the output is suitable for morphological operations, which rely on binary inputs to effectively clean up and refine the detected events or features. These preprocessing steps are necessary due to the unique characteristics of event-based cameras. Unlike frame-based cameras, where one must first separate the moving insect from a static background, event-based cameras inherently record only changes in the scene. As a result, tracking a honey bee’s movement involves directly following its motion signals rather than performing explicit background modeling.
\paragraph{Morphological Operations - Erosion and Dilation}
Morphological operations are image processing techniques that modify the structure of objects in a binary image to improve their clarity. In this context, we employ two fundamental operations: erosion and dilation.
Erosion is used to remove small-scale noise that conceals the features of interest. It takes a structuring element as an argument. This binary matrix defines the neighborhood around each pixel, slides over the image, and sets the pixel in the output image to the minimum value of the input pixels covered by the structuring element. By shrinking foreground objects' boundaries, erosion can simplify the image and remove irrelevant details, making the significant structures more distinguishable, like the bodies of honeybees.

Dilation is a complementary operation to erosion, which dilates the boundary of foreground objects. Like erosion, it also uses a structuring element; however, it sets the pixel in the output image to the maximum of all the input pixels covered by the structuring element. Therefore, this operation will accentuate the related structures within the image so that the honey bees are more apparent and more accessible to detect. Dilation strengthens the connectivity of structures and reestablishes parts that might have been broken off by erosion.

\paragraph{Contouring}
Following preliminary morphological operations, contouring techniques are applied directly to the preprocessed images from event-based cameras. Contours are detected using OpenCV \citep{bradski2000}, which is defined as continuous curves delineating points of uniform intensity. It is also a function that returns contours and gives information about their hierarchy, making it possible to differentiate external contours from internal ones, enabling more subtle analysis and filtering. This reduces computational burdens and increases processing efficiency. Further contour-based features, like areas, perimeters, centroids, and bounding boxes, are computed for visualization to differentiate honeybees from all other entities in the image. Finally, size, shape, and aspect ratio filtering of the contours remove unsuitable objects, sharpening the detection precision.

\paragraph{Rectification and Stereo Matching}
In stereo vision, the correspondence problem is considered a central challenge: determining which features in the left image correspond to the same physical points in the right image. Solving this problem will ensure that every matched pair of points corresponds to the same object in the scene from two different perspectives, which becomes very important to estimate depth and perform 3D reconstruction more accurately \citep{Mahadeeswara}. However, establishing these correspondences from images might be complex and prone to errors because of variable viewing angles, lighting conditions, and possible occlusions.
Our method finds matching points between two images by first adjusting them so that any corresponding features appear along the same horizontal lines. This process, called image rectification, simplifies the search for matches because we only have to look along these straight, horizontal "epipolar" lines rather than searching all over the image. This reduces the number of dimensions in the search space for correspondence. Because of image rectification, all possible matches lie on the same horizontal line in both images, reducing two-dimensional search space to one dimension. The cv2.remap function from the OpenCV library \citep{bradski2000} was used to apply pixel transformations to rectify the stereo images.
Following rectification, the fundamental matrix F, which encodes the geometric relationship between two images of the same scene, taken from different viewpoints, is used to compute the epipolar line in one image for each feature point detected in the other. The epipolar line in Fig. 3(d) represents all possible locations for the corresponding point in the second image based on the known geometry of the stereo camera configuration. The perpendicular distance between each candidate feature in the second image and the corresponding epipolar line from the first image is calculated to identify the optimal match. A smaller perpendicular distance indicates a higher likelihood of correspondence, as it implies proximity to the theoretical epipolar constraint.
This matching process is further refined by applying a cost-based optimization using the Hungarian algorithm, also known as the Kuhn-Munkres algorithm \citep{kuhn1955hungarian, munkres1957algorithms}, to minimize the cumulative epipolar distance across all potential matches. This ensures each feature in one image corresponds uniquely to a feature in the other, based on minimal cumulative distance. For cases when there is high density of honeybees and several honeybees  are detected on the same horizontal epipolar line, the cost function includes additional parameters: distance, velocity, and size. These allow further differentiation to ensure that the motion history and size consistency across each bee are considered. For example, if there is an abrupt change in the size or great leap in the velocity, it is rejected as false correspondence. This method incorporates several parameters so that, when the cost is below a pre-set threshold, it allows the algorithm to identify object matches from frame to frame and achieve more accurate correspondence in those multiple-individual scenarios that might otherwise confuse a matching algorithm. 
Once correspondences have been established, the disparity can be calculated as the horizontal distance between the matched points in rectified images. This then gives ground for triangulation-based depth estimation, enabling the calculation of 3D coordinates for the matched features. The method provides consistent features over frames by assigning a unique ID to every feature matched successfully, ensuring reliable 3D reconstruction of a scene along with depth-based analyses.

\subsubsection{Tracking Phase}

Our multi-object tracking method uses a Kalman Filter \citep{li2010} to track the trajectories of honeybees. This process is conducted in 2D for each camera perspective and subsequently extended to 3D space for a comprehensive analysis. Every detected and matched object is identified by a unique identifier, and its position, size, and velocity are precisely recorded. With new frames being processed, the algorithm uses its current velocity and position to predict the next likely position of each individual honeybee it is tracking. This prediction makes the continuity of track possible to be retained even when a bee briefly leaves the field of vision or gets occluded by others. These predicted positions are then updated when new measurements from stereo vision arrive. This method is based on comparing the predicted and actual positions, thus minimizing the error and refining the estimated trajectory for a honeybee.
Dynamic management of object states is an integral part of the tracking process as each trajectory is constantly checked for consistency, and those that deviate significantly from usual behavior are flagged. This ensures that tracking is reliable, yet able to adapt to erratic movements often exhibited by small and fast-moving bees. Moreover, the entrance of new bees into the scene is detected and integrated smoothly into the system, and bee ID's that have not been detected within a certain period are removed. Our tracking approach also provides a mechanism to handle missed detections or even to reidentify objects that have been transiently lost. We have implemented a parameter that defines the maximum number of lost frames for every tracker. If a bee ID is not detected for a certain number of frames around it's position, then it will be considered a lost bee, and a new tracker is set for every new-coming bee into the view. On the other hand, if a previously tracked bee re-enters its former position within the defined temporal and spatial thresholds, it will be recognized as the same individual, enabling continuous tracking. Since event-based cameras do not record static objects, a hovering bee generates no new events. Therefore, incorporating a wait-and-see mechanism becomes essential: the system must temporarily hold the bee’s last known position until it resumes movement and triggers new events, ensuring that the bee’s trajectory can be followed seamlessly. The method herein will ensure that each individual bee's trajectories are identified correctly and kept on continuous tracks.

\section{Results and Discussion}

We employed event cameras in a stereoscopic configuration to track the flight activity of bees around the hive entrance. The cameras were positioned 3.4 meters from the edge of the hive, providing a field of view of approximately 2.5 meters along the X-axis—sufficient to capture both the insects’ approach and exit behavior. Over a 5-minute recording period, a total of 652 bees were identified and tracked. The reconstructed trajectories, shown in Figure 4, confirm that the hive entrance was triangulated correctly, as flight paths converge at an image depth of about 3.4 meters. This setup demonstrates the viability of event-based stereo vision for monitoring rapid insect movements under real-world conditions.

Building on these tracked trajectories, we investigated the tortuosity of bee flight paths. Tortuosity is defined as the ratio of the actual path travelled to the straight-line distance between the starting point and destination (in this case, the hive). Among the 652 tracked bees, those that traveled from their first detection point all the way to the hive covered an approximate straight-line distance of 2.5 meters. However, their actual flight paths often exceeded this distance. On average, bees travelled about 5.9 meters and exhibited an average tortuosity of 2.35, with a standard deviation of 0.48—indicating that they travelled more than twice the straight-line distance. Temporary factors, such as gusty winds and heavy traffic near the hive entrance, can cause these values to rise slightly by forcing bees to navigate more complex paths to maintain orientation and stability.

To further emphasize the efficiency of event cameras, we compared the data volume generated by our event-based system with that of a GoPro Hero 10 \citep{gopro2023}, both operating over the same 5-minute interval. The event camera produced 151 MB of data, whereas the GoPro created a 1.7 GB video file. This comparison underscores a roughly 91\% reduction in file size with event cameras, highlighting their potential for long-term, storage-efficient monitoring of fast-moving subjects such as pollinators, agricultural pests, and insects with evasive flight behaviors like dragonflies or wasps.

All recordings were processed on an HP Pavilion laptop running an Intel Core i5, 9th Generation, equipped with a 4 GB GTX 1650Ti graphics card. This moderate-specification setup easily handled the asynchronous event streams, removing the need to convert raw data into conventional video formats. As a result, event cameras open new avenues for real-time insect behavior analysis, especially when paired with machine learning models. Such approaches enable higher predictive accuracy for detecting ecologically relevant events—ranging from seasonal pollinator patterns to shifts in insect distribution driven by changing environmental conditions.

\begin{figure*}[!hb]
\begin{center}
\includegraphics[height=12cm]{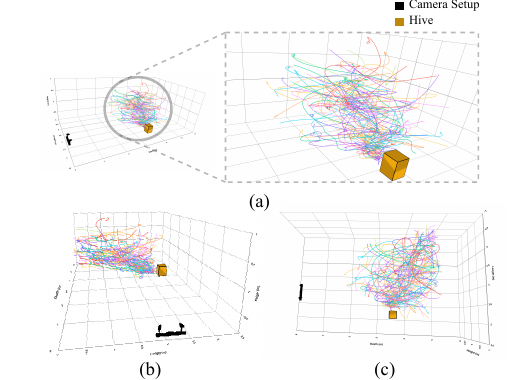}    
\caption{\textbf{Illustration of honeybee trajectories observed in a 40-second recording session:} (a) Displays all trajectories of honeybees during a session. There are two visualizations: the full axis plot, including the depth data for all the movements of bees in relationship to the hive (yellow box) and; a magnified hive entrance view capturing highly specific detail around the hive. (b) a corresponding side view that contextualizes the relative view of the camera with trajectories of bees, with altitude data. (c) A top-down view of the reconstructed trajectories, providing an overview of movement patterns.}

\label{fig:fig3}                                 
\end{center}                                 
\end{figure*}

\subsection{Limitations and Future Directions for Improvement}
There are some important limiting factors to consider when using this method. One such major limitation is that the cameras only capture changes in brightness intensity at the pixel level in positive and negative polarities. Hence, detailed information regarding the appearance is not obtained, which includes potentially crucial aspects such as color, or other distinctive features of the insects, see  Fig.\ref{fig:biassettings} and SUPP Fig. S2(a). Balancing the different tuning parameters of the cameras is required to ensure the only relevant movements are registered. However in this process a tradeoff between resolving finer features on the insect body versus minimizing background noise will be needed. Using this approach outlines or silhouettes of the moving insects are well captured. As much as the outline can be beneficial in recognizing general movement patterns, it becomes less intuitive to the end users, who base their interpretations on easily recognizable visual information. Detailed morphological features can be captured if the subject is non-homogeneous and occupies a large portion of the field of view which would be unlikely when tracking freely moving insects in the field.  

Another important limitation is that the static background information is also lost, which can be of valuable importance in behavioral studies. For instance, when studying how ants navigate their environment or the search behavior of wasps near a hive, the loss of background information hampers the ability to contextualize their behavior in relation to their environment. Moreover, event cameras cannot calculate the exact orientations of insects since they do not provide any structural information. The lack of detailed information about the features implies that orientation must be estimated purely based on position and silhouette shape, which is ambiguous and will naturally yield reduced accuracy. 

Such limitations point to the need for further sophisticated imaging and analysis techniques. One potential approach would be to synchronize event and frame camera setups that will truly unlock the potential of event cameras for tracking and analyzing insect motions while retaining the background information through the standard imaging system \citep{gebauer2024}. Such a hybrid setup would allow dynamic motion detected by event cameras to include comprehensive visual and structural information from frame cameras. This combination would enhance the study of complex behaviors and facilitate more accurate species identification and behavioral analysis, leveraging the strengths of both types of cameras to provide a more comprehensive view of animal and insect dynamics. 

Another limitation is the lack of data processing tools for asynchronous event streams and the need to convert data streams to frames to apply existing image processing tools. Efficiently processing events based on their timing and clustering would be one approach for real-time subject identification and tracking. Machine learning models that process event clusters for insect identification would also help extract further information.


\ack{We extend our thanks to Mr. Pugazhendhi Priyan for his contribution in creating the Blender illustration of the calibration setup (Fig. \ref{fig:fig3} (a)). We are grateful to Mr. Saurabh Chaughule for contributing his time and expertise in designing the CAD model for the camera setup and 3D printing the camera mounts. Additionally, we appreciate Mr. Yuvan Kamalakanthan for his assistance with videography, which was crucial in clearly explaining the methods and techniques used.}

\competing{The authors declare no competing interests.}

\contribution{P.S.: conceptualization, developing methodology and software, collecting data, writing manuscript.
M.L.: conceptualization, developing methodology and software. J.H.:  developing methodology and writing manuscript. M.M: developing methodology and writing manuscript. M.S.: conceptualization, developing methodology and writing manuscript. S.R.: conceptualization, developing methodology and software and writing manuscript. 
}

\bibliographystyle{unsrt}
\bibliography{references}

@article{hagler2001,
  title={Methods for marking insects: current techniques and future prospects},
  author={Hagler, James R and Jackson, Charles G},
  journal={Annual review of entomology},
  volume={46},
  number={1},
  pages={511--543},
  year={2001},
  publisher={Annual Reviews 4139 El Camino Way, PO Box 10139, Palo Alto, CA 94303-0139, USA}
}

@article{sendova,
  title={Task allocation in ant colonies within variable environments (a study of temporal polyethism: experimental)},
  author={Sendova-Franks, Ana and Franks, Nigel R},
  journal={Bulletin of Mathematical Biology},
  volume={55},
  pages={75--96},
  year={1993},
  publisher={Springer}
}

@article{boiteau2001,
  title={Electronic tags for the tracking of insects in flight: effect of weight on flight performance of adult Colorado potato beetles},
  author={Boiteau, Gilles and Colpitts, Bruce},
  journal={Entomologia experimentalis et applicata},
  volume={100},
  number={2},
  pages={187--193},
  year={2001},
  publisher={Wiley Online Library}
}

@article{moreau2011,
  title={Use of radio-tagging to map spatial organization and social interactions in insects},
  author={Moreau, Mathieu and Arrufat, Patrick and Latil, G{\'e}rard and Jeanson, Rapha{\"e}l},
  journal={Journal of Experimental Biology},
  volume={214},
  number={1},
  pages={17--21},
  year={2011},
  publisher={Company of Biologists}
}

@article{naqvi2022,
  title={Camera traps are an effective tool for monitoring insect--plant interactions},
  author={Naqvi, Qaim and Wolff, Patrick J and Molano-Flores, Brenda and Sperry, Jinelle H},
  journal={Ecology and Evolution},
  volume={12},
  number={6},
  pages={e8962},
  year={2022},
  publisher={Wiley Online Library}
}

@article{droissart2021,
  title={PICT: A low-cost, modular, open-source camera trap system to study plant--insect interactions},
  author={Droissart, Vincent and Azandi, Laura and Onguene, Eric Rostand and Savignac, Marie and Smith, Thomas B and Deblauwe, Vincent},
  journal={Methods in Ecology and Evolution},
  volume={12},
  number={8},
  pages={1389--1396},
  year={2021},
  publisher={Wiley Online Library}
}

@article{straw2011,
  title={Multi-camera real-time three-dimensional tracking of multiple flying animals},
  author={Straw, Andrew D and Branson, Kristin and Neumann, Titus R and Dickinson, Michael H},
  journal={Journal of The Royal Society Interface},
  volume={8},
  number={56},
  pages={395--409},
  year={2011},
  publisher={The Royal Society}
}

@article{sakakibara2012,
  title={Note: High-speed optical tracking of a flying insect},
  author={Sakakibara, Jun and Kita, Junichiro and Osato, Naoyuki},
  journal={Review of scientific instruments},
  volume={83},
  number={3},
  year={2012},
  publisher={AIP Publishing}
}

@inproceedings{gebauer2024,
  title={Towards a Dynamic Vision Sensor-based Insect Camera Trap},
  author={Gebauer, Eike and Thiele, Sebastian and Ouvrard, Pierre and Sicard, Adrien and Risse, Benjamin},
  booktitle={Proceedings of the IEEE/CVF Winter Conference on Applications of Computer Vision},
  pages={7157--7166},
  year={2024}
}

@misc{prophesee2024,
  author    = {Prophesee},
  title     = {Event Camera Evaluation Kit 4 HD IMX636},
  year      = {2022},
  note      = {Retrieved from \url{https://www.prophesee.ai/event-camera-evk4/}}
}

@article{bradski2000,
  title={OpenCV},
  author={Bradski, Gary and Kaehler, Adrian and others},
  journal={Dr. Dobb’s journal of software tools},
  volume={3},
  number={2},
  year={2000}
}

@inproceedings{li2010,
  title={A multiple object tracking method using Kalman filter},
  author={Li, Xin and Wang, Kejun and Wang, Wei and Li, Yang},
  booktitle={The 2010 IEEE international conference on information and automation},
  pages={1862--1866},
  year={2010},
  organization={IEEE}
}

@misc{gopro2023,
  author    = {GoPro},
  title     = {HERO10 Black Announcement},
  year      = {2021},
  note      = {Retrieved from \url{https://gopro.com/en/us/news/hero10-black-announce}}
}

@article{kuhn1955hungarian,
  title={The Hungarian method for the assignment problem},
  author={Kuhn, Harold W},
  journal={Naval research logistics quarterly},
  volume={2},
  number={1-2},
  pages={83--97},
  year={1955},
  publisher={Wiley Online Library}
}

@article{munkres1957algorithms,
  title={Algorithms for the assignment and transportation problems},
  author={Munkres, James},
  journal={Journal of the society for industrial and applied mathematics},
  volume={5},
  number={1},
  pages={32--38},
  year={1957},
  publisher={SIAM}
}

@article{zhang2000camera,
  title={A flexible new technique for camera calibration},
  author={Zhang, Zhengyou},
  journal={IEEE Transactions on pattern analysis and machine intelligence},
  volume={22},
  number={11},
  pages={1330--1334},
  year={2000},
  publisher={IEEE}
}

@article {Mahadeeswara,
	author = {Mahadeeswara, Mandiyam Yadav and Srinivasan, Mandyam V},
	title = {Reconstruction of Three-Dimensional Trajectories of Honeybees Flying in High-Density Aerial Environments},
	elocation-id = {2022.11.24.517808},
	year = {2022},
	doi = {10.1101/2022.11.24.517808},
	publisher = {Cold Spring Harbor Laboratory},
	URL = {https://www.biorxiv.org/content/early/2022/11/24/2022.11.24.517808},
	eprint = {https://www.biorxiv.org/content/early/2022/11/24/2022.11.24.517808.full.pdf},
	journal = {bioRxiv}
}

@INPROCEEDINGS{ZouLing,
  author={Zou, Ling and Li, Yan},
  booktitle={2010 International Conference on Audio, Language and Image Processing}, 
  title={A method of stereo vision matching based on OpenCV}, 
  year={2010},
  volume={},
  number={},
  pages={185-190},
  doi={10.1109/ICALIP.2010.5684978}}

@article{capaldi2000ontogeny,
  title={Ontogeny of orientation flight in the honeybee revealed by harmonic radar},
  author={Capaldi, Elizabeth A and Smith, Alan D and Osborne, Juliet L and Fahrbach, Susan E and Farris, Sarah M and Reynolds, Donald R and Edwards, Ann S and Martin, Andrew and Robinson, Gene E and Poppy, Guy M and others},
  journal={Nature},
  volume={403},
  number={6769},
  pages={537--540},
  year={2000},
  publisher={Nature Publishing Group UK London}
}

@article{colpitts2004harmonic,
  title={Harmonic radar transceiver design: Miniature tags for insect tracking},
  author={Colpitts, Bruce G and Boiteau, Gilles},
  journal={IEEE Transactions on Antennas and Propagation},
  volume={52},
  number={11},
  pages={2825--2832},
  year={2004},
  publisher={IEEE}
}

\end{document}